\documentclass{article}

\usepackage[preprint]{timeseries_workshop}

\usepackage[utf8]{inputenc} % allow utf-8 input
\usepackage[T1]{fontenc}    % use 8-bit T1 fonts
\usepackage{hyperref}       % hyperlinks
\usepackage{url}            % simple URL typesetting
\usepackage{booktabs}       % professional-quality tables
\usepackage{amsfonts}       % blackboard math symbols
\usepackage{nicefrac}       % compact symbols for 1/2, etc.
\usepackage{microtype}      % microtypography
\usepackage{xcolor}         % colors
\usepackage{amsmath}
\usepackage{graphicx}
\setcitestyle{round}
\usepackage{amssymb}
\usepackage{algorithm}
\usepackage{algorithmic}
\usepackage{subfigure}
\usepackage{amsfonts}

\title{Analyzing Patient Daily Movement Behavior Dynamics Using Two-Stage Encoding Model}

\author{%
  Jin Cui\thanks{Corresponding author: \texttt{jc9223@ic.ac.uk}. Jin Cui and Alexander Capstick contributed equally.} \\
  Imperial College London\\
  UK Dementia Research Institute, Care Research and Technology Centre
  \And
  Alexander Capstick\footnotemark[1] \\
  Imperial College London\\
  UK Dementia Research Institute, Care Research and Technology Centre
  \And
  Payam Barnaghi\thanks{Payam Barnaghi and Gregory Scott contributed equally.} \\
  Imperial College London\\
  UK Dementia Research Institute, Care Research and Technology Centre
  \And
  Gregory Scott\footnotemark[2] \\
  Imperial College London\\
  UK Dementia Research Institute, Care Research and Technology Centre
}

\begin{document}

\maketitle

\begin{abstract}
  In the analysis of remote healthcare monitoring data, time series representation learning offers substantial value in uncovering deeper patterns of patient behavior, especially given the fine temporal granularity of the data. In this study, we focus on a dataset of home activity records from people living with Dementia. We propose a two-stage self-supervised learning approach. The first stage involves converting time-series activities into text strings, which are then encoded by a fine-tuned language model. In the second stage, these time-series vectors are bi-dimensionalized for applying PageRank method, to analyze latent state transitions to quantitatively assess participants behavioral patterns and identify activity biases. These insights, combined with diagnostic data, aim to support personalized care interventions.
\end{abstract}

\section{Introduction}

In remote healthcare monitoring applications, the use of wearables and Internet of Things (IoT) devices to continuously collect time-series data, often with second-level accuracy or finer, has become increasingly common. However, the sheer scale of such data makes it difficult for human experts to analyze or use directly, necessitating the use of time-series deep learning techniques for effective analysis and diagnosis.

Training on large volumes of unlabeled time-series data poses a significant challenge. Semi-supervised and unsupervised methods are typically employed to encode and extract data features for downstream tasks like classification or regression, demonstrating their ability to capture deep features. Semi-supervised methods, such as nearest neighbor contrastive learning and temporal relation prediction, efficiently utilize both labeled and unlabeled data, improving the quality of representations for downstream tasks like classification \citep{kim_semi-supervised_2024,fan_semi-supervised_2021}. Unsupervised methods focus on learning robust representations without relying on labels, often leveraging contrastive learning techniques and innovative data augmentations to capture key temporal patterns \citep{franceschi_unsupervised_2019,lee_spatio-temporal_2024}. Attention mechanisms and domain-adaptive techniques further enhance the interpretability of encoded features, aligning them more closely with human intuition and domain-specific insights \citep{lyu_improving_2018}. However, this strategy faces two challenges: first, labeling criteria for time-series data is often vague, which can significantly impact model performance; second, the encoded data remain vast, unintuitive, and difficult to interpret \citep{ye_lbp4mts_2023,hill_semi-supervised_2022}.

In this work, we focus on time-series data characterized by irregular discrete values. Extending the methods introduced in \citet{capstick2024representation}, we present preliminary results of a second-order representation learning method designed to aid in clustering, identifying similar clinical cases, and uncover patients' interpretable behavioral patterns. This is achieved through a large language model encoding combined with a two-dimensional vectors representation and transfer pattern analysis.

\subsection{Background}

Time-series forecasting is primarily to predict future values based on previously observed data points. Traditional statistical methods, most notably the Autoregressive Integrated Moving Average (ARIMA) model, have long been utilized due to their mathematical simplicity and flexibility in application \citep{rizvi_arima_2024,kontopoulou_review_2023}. While ARIMA remains a staple for scenarios where data exhibits linear patterns, recent developments in machine learning have introduced sophisticated models capable of capturing non-linear dependencies, thus offering potential improvements in forecasting accuracy and robustness \citep{masini_machine_2023,rhanoui_forecasting_2019}.

The advent of the Generative Pre-trained Transformer (GPT) by OpenAI marked a significant milestone in the field of natural language processing \citep{brown_language_2020}, catalyzing a wave of innovations in large language models (LLMs). Large Language Models (LLMs) have profoundly transformed natural language processing and are increasingly being considered for diverse applications beyond text, such as time series data analysis. The study by \citep{bian_multi-patch_2024} presents a framework that adapts LLMs for time-series representation learning by conceiving time-series forecasting as a multi-patch prediction task, introducing a patch-wise decoding layer that enhances temporal sequence learning. Similarly, \citep{liu_autotimes_2024} propose a model which leverages the autoregressive capabilities of LLMs for time series forecasting. In \citet{capstick2024representation}, the authors apply a GPT-based text encoder to string representations of in-home activity data to enable vector searching and clustering. Using a secondary modelling stage, we extend these ideas to enable further analysis and interpretability.

PageRank, originally developed to rank web pages, is an algorithm designed to assess the importance of nodes within a directed graph by analyzing the structure of links within networks \citep{page_pagerank_1999}. While it was initially created for search engines, its application has since expanded across various disciplines. For instance, in biological networks, \citep{ivan_when_2011} employed personalized PageRank to analyze protein interaction networks, providing scalable and robust techniques for interpreting complex biological data. Similarly, \citep{banky_equal_2013} introduced an innovative adaptation of PageRank for metabolic graphs. This cross-disciplinary application of PageRank highlights its potential for analyzing complex systems beyond its original domain.

\subsection{Our Contribution}

We propose an integrated approach for discovering latent states of activity. This method comprises several key steps:

\begin{enumerate}
    \item \textbf{Temporal Data Preprocessing}: The raw temporal data is first preprocessed to remove noise and standardize the data for consistency.
    \item \textbf{Language Model Encoding}: A language model is trained on our dataset to encode the preprocessed temporal data into high-dimensional vector representations. To enhance the model's learning capability, we perform pseudo labeling using one-hot similarity. This allows the model to better capture temporal dependencies and patterns in the data.
    \item \textbf{Dimensionality Reduction and Clustering}: To visualize the high-dimensional embeddings, we apply dimensionality reduction techniques such as t-SNE to project the data into a 2D space. Clustering algorithms are then used to identify distinct latent states within the data.
    \item \textbf{Transition Pattern Analysis}: By defining a transition matrix between different latent states, we apply the PageRank algorithm to analyze the transition patterns. This allows us to determine the importance or influence of each state in the transition graph, providing insights into both the state dynamics and patient behavior patterns.
\end{enumerate}

This analytical framework will aid in the clinical diagnosis of patients and support the development of personalized care programs. Appendix \ref{append_data_reproduction}
discusses the availability of dataset and code for this work.

\section{Methods}
\label{method}

\subsection{Mathematical Foundations of the Model}
Given a discrete data sample $\mathbf{X} = \{x_1, x_2, \dots, x_n\}$, the following steps describe the transformation process:

1. \textbf{Sampling and Text Conversion:} Each sample $x_i$ is converted into a text representation $T(x_i)$.

2. \textbf{Language Model Encoding:} A pre-trained language model $f_{\text{LM}}$ is applied to obtain high-dimensional vector embeddings for the text data:
\[
\mathbf{h}_i = f_{\text{LM}}(T(x_i)), \quad \mathbf{h}_i \in \mathbb{R}^d.
\]

3. \textbf{Dimensionality Reduction:} The high-dimensional embeddings are projected into a 2D space using a dimensionality reduction method $\Phi$, such as t-SNE:
\[
\mathbf{z}_i = \Phi(\mathbf{h}_i), \quad \mathbf{z}_i \in \mathbb{R}^2.
\]

4. \textbf{PageRank and Deep State Vector Extraction:} A transition matrix $\mathbf{P}$ between points in 2D space is constructed, and the PageRank algorithm is applied to further reduce the dimensionality:
\[
\mathbf{v}_i = \text{PageRank}(\mathbf{P}), \quad \mathbf{v}_i \in \mathbb{R}^k, \quad k \ll d.
\]

The final low-dimensional vectors $\mathbf{v}_i$ capture deep semantic relationships from the original data.

\subsection{The Dataset}

We obtained a dataset collected from 134 people diagnosed with dementia, capturing their home location movement data between July 1, 2021, and January 30, 2024. The dataset records the time entering different rooms and sleeping mats, alongside clinical metrics such as MMSE \citep{kurlowicz_mini-mental_1999}, ADAS-Cog \citep{kueper_alzheimers_nodate} scores from regular tests. It also includes details on various factors such as demographic data, comorbidities, and other medical information. The dataset contains a total of 66,096 recording days. A more detailed description of the dataset is provided in Appendix \ref{append_dataset}. After excluding patients with missing data, the final dataset used for further analysis contained 50 participants with complete information.

\subsection{Our Framework}

\begin{figure}[htbp]
    \centering
    \includegraphics[width=0.8\textwidth]{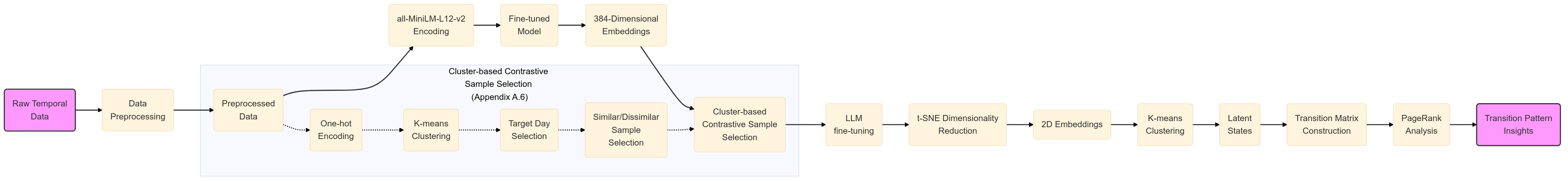}
    \caption{Flowchart of the framework.}
    \label{fig:flowchart-diagram}
\end{figure}

\begin{figure}[htbp]
    \centering
    \includegraphics[width=0.8\textwidth]{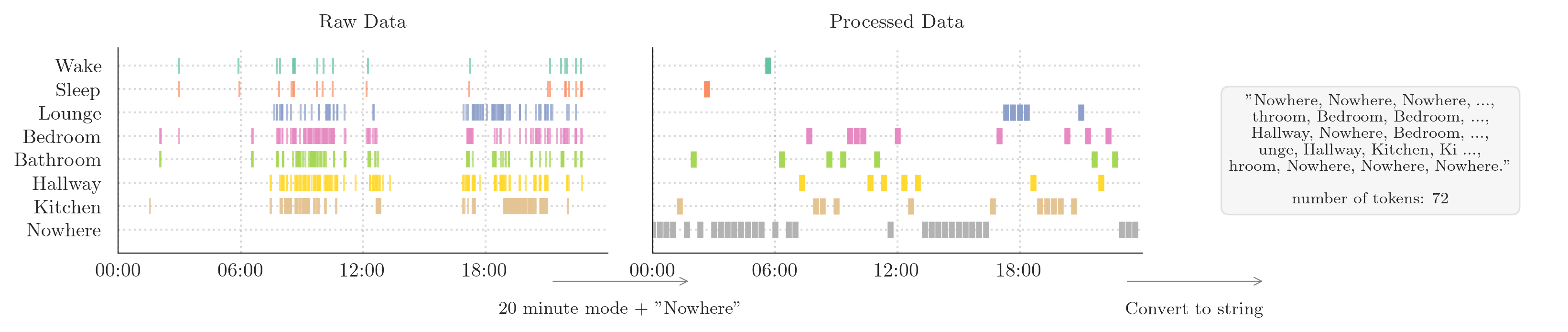}
    \caption{Flowchart of data preprocessing. The figure illustrates the monitoring data for a single participant over the course of one day. The left graph displays the raw, unprocessed measurements. In the middle graph, the data is rectified into 20-minute intervals, where periods of inactivity are labeled as "nowhere." Within each window, the most frequent location, excluding "nowhere," is identified and recorded. The right graph presents the corresponding text strings, which are formatted for interpretation by the language model.}
    \label{fig:preprocess-diagram}
\end{figure}

Our framework consists of several key stages: data preprocessing and encoding, latent state discovery, and transition pattern analysis, as shown in Figure \ref{fig:flowchart-diagram}. First, we preprocess the raw temporal data to remove noise and ensure consistency. This process is illustrated in Figure \ref{fig:preprocess-diagram}. We then utilize the all-MiniLM-L12-v2 model \citep{muennighoff_mteb_2023} as the language model encoder. This model excels at capturing similarities in textual information, making it suitable for analyzing similarities between recorded dates and uncovering potential relationships. We fine-tune the model using its pretrained weights to adapt to the specific characteristics of our dataset. The preprocessed temporal data is then encoded into 384-dimensional vector representations, capturing the inherent temporal dependencies and patterns within the data. Given the unlabeled nature of our temporal data, we adopt a Cluster-based Contrastive Sample Selection method and triplet loss for training and evaluation. Further details on the language model training process are described in Appendix \ref{append_training}. To visualize and interpret the high-dimensional embeddings, we apply the t-SNE dimensionality reduction technique \citep{van_der_maaten_viualizing_2008} to project the data into a 2D space. K-means clustering is then employed to identify distinct latent states within the data. As the data points are temporally ordered, this 2D map allows us to visualize each participant's latent activity map as their movement pattern projected onto a specific dimension. Finally, we define a transition matrix between the different latent states and apply the PageRank algorithm \citep{page_pagerank_1999} to analyze the transition patterns, details of this algorithm are available in Appendix \ref{pagerank}. This analysis provides insights into the flow and importance of each latent state within the overall temporal dynamics of the data, .

\section{Experiments}
\label{experiment}

After clustering the text vectors of the test set using K-means, we identified the optimal clustering result at 5 clusters (Appendix \ref{tsneplot1}), suggesting five latent states across all single-day, single-participant behavioral patterns. Figure \ref{tsneplotpic1} shows the clustering outcomes following dimension reduction of the embeddings using t-SNE. By examining the transformation of individual vectors in two dimensions, we can visualize the behavioral trajectories of different participants within the embedding space (see Appendix \ref{tsneplot2} for more participant visualizations). Collaborating with clinical experts, we can explore the semantics represented by these clusters and their relationship to patient medical characteristics.

By integrating other participant data -- such as housing type and whether they live alone, we can begin to infer participant behavior and care needs.

More significantly, by applying the random wandering model and the PageRank algorithm to these two-dimensional plots, in combination with clinical expert opinions and diagnostic results, we can quantitatively assess the deeper semantics represented by the vector clusters, or latent states. By reducing the complex vector matrix into a simplified (1,5) vector(Appendix \ref{pagerank_vis}), we can explore semantic characteristics to each of the dimensions. For example, once a unique deep vector is generated for each participant, we can easily identify the disease type, age, MMSE, ADAS-Cog scores, and their rate of change for the three patients most and least similar to any given participant. This revealed that the clinical differences between similar groups were indeed smaller in features like age, change in ADAS-Cog score (Appendix \ref{similar}). Furthermore, clustering the latent vectors of all 50 participants highlighted pronounced differences between clusters, each offering potential explanations for the link between activity data and clinical diagnosis (Appendix \ref{clustering}).

Looking forward, now that we have established a process for encoding deep vectors, we could explore transforming this approach into a generative model. Such a model could be used to generate sensitive and hard-to-obtain medical datasets for purposes like data augmentation or alignment, a strategy proven effective in the training of large language models\citep{li_synthetic_2023}.

\section{Conclusion}

In conclusion, our initial results demonstrate that by applying our framework, we show that our latent states vector based on patients daily activity patterns can be useful for exploring behavior dynamics. While these findings offer a promising approach to exploring the relationship between behavior and clinical characteristics, further research is needed to refine the model and validate its broader applications, including potential use in medical data augmentation.

\bibliography{references,bib_2}

%%%%%%%%%%%%%%%%%%%%%%%%%%%%%%%%%%%%%%%%%%%%%%%%%%%%%%%%%%%%

%%%%%%%%%%%%%%%%%%%%%%%%%%%%%%%%%%%%%%%%%%%%%%%%%%%%%%%%%%%%

\appendix
\section{Appendix}

\subsection{Availability of datasets and code}
\label{append_data_reproduction}

The IPython notebooks used to build this framework will be released after review. The dataset and IPython notebooks used to plot the data will not be made public due to their sensitivity. The experiments were conducted using Python 3.11.5, Torch 2.4.0 \citep{ansel_pytorch_2024}, Transformers 4.44.2 \citep{wolf_transformers_2020}, Sentence-Transformers 2.7.0 \citep{pedregosa_scikit-learn_2011}, Scikit-Learn 1.3.2 \citep{pedregosa_scikit-learn_2011}, NumPy 1.26.4 \citep{harris_array_2020}, SciPy 1.13.1 \citep{virtanen_scipy_2020} and Pandas 2.1.4 \citep{mckinney_data_2010}. 

\subsection{Detailed Description of The Dataset}
\label{append_dataset}

The dataset used for in-home activity monitoring was collected via passive infrared sensors installed at multiple locations in the homes of individuals with dementia, along with sleep pads placed under their mattresses. These infrared sensors detect motion within a range of up to nine meters, at a maximum angle of forty-five degrees diagonally upward. Sensors were placed in lounges, kitchens, hallways, bedrooms, and bathrooms, allowing for detailed tracking of participants' movements within and between these areas.

\begin{figure}[htbp]
    \centering
    \includegraphics[width=0.8\textwidth]{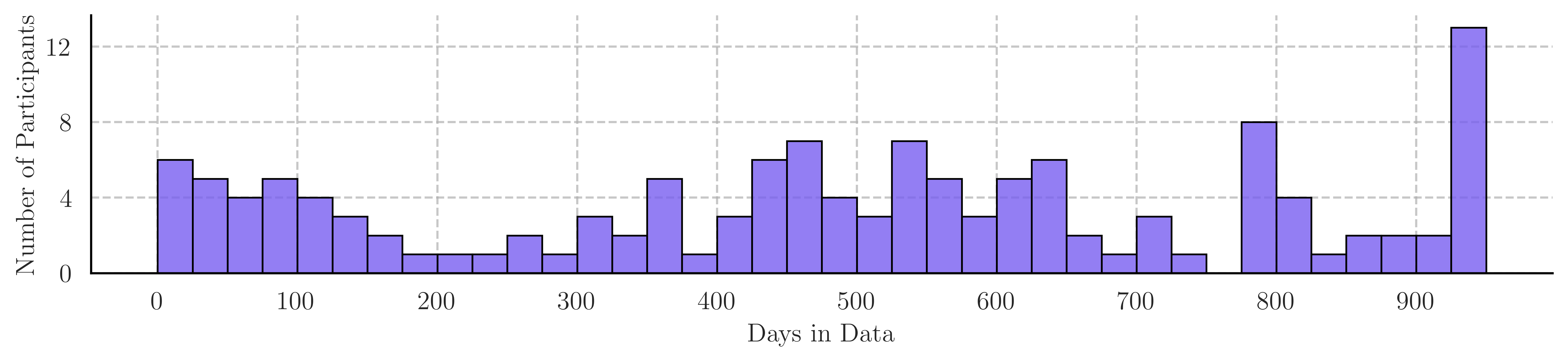}
    \caption{Daily histogram.}
    \label{histo}
\end{figure}

\begin{figure}[htbp]
    \centering
    \includegraphics[width=0.8\textwidth]{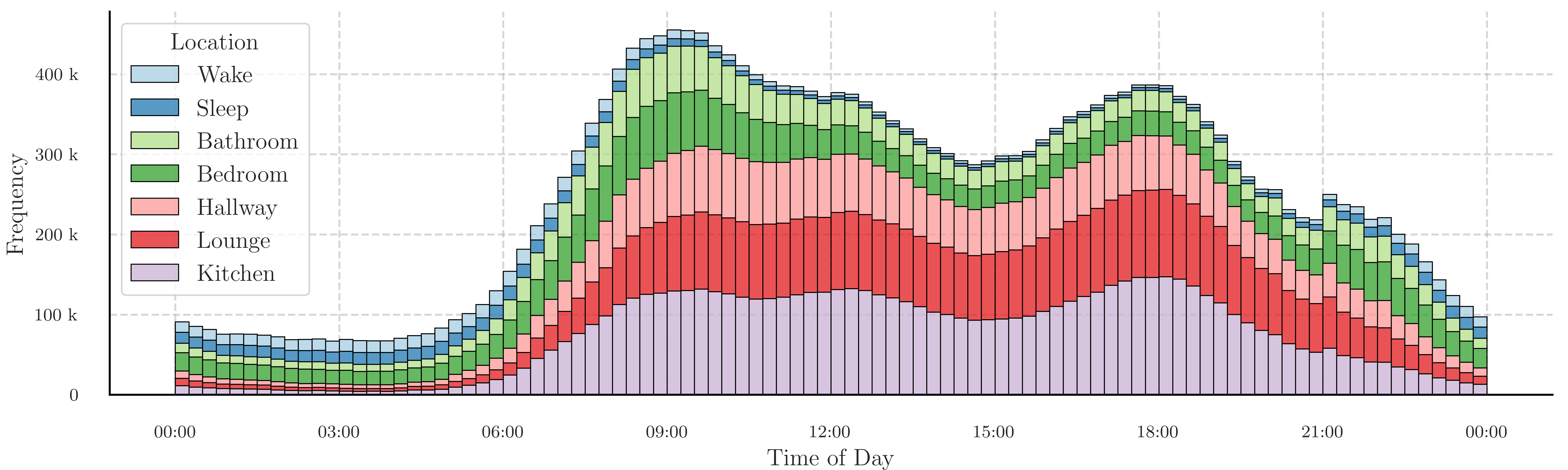}
    \caption{Location Histogram.}
\label{location}
\end{figure}

\begin{figure}[htbp]
    \centering
    \includegraphics[width=0.8\textwidth]{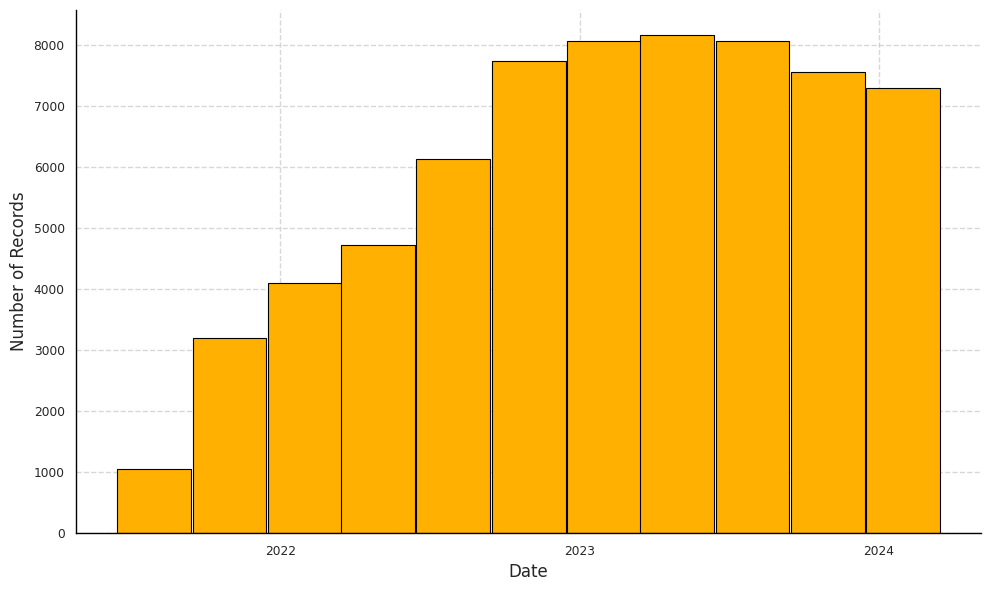}
    \caption{Three month distribution.}
\label{threemonth}
\end{figure}

We analyzed data recorded between July 1, 2021, and January 30, 2024, amounting to 66,096 participant-days for 134 individuals. Figure \ref{histo} illustrates the distribution of logged days per participant. Each data point includes the participant ID, a timestamp (accurate to the second), the location of detected activity, or sleep pad data indicating whether the participant entered or left their bed. The dataset contains a total of 24,467,307 individual records, as depicted in Figure \ref{location}. Figure \ref{threemonth} shows the three month interval distribution of records.

\begin{figure}[htbp]
    \centering
    \includegraphics[width=0.8\textwidth]{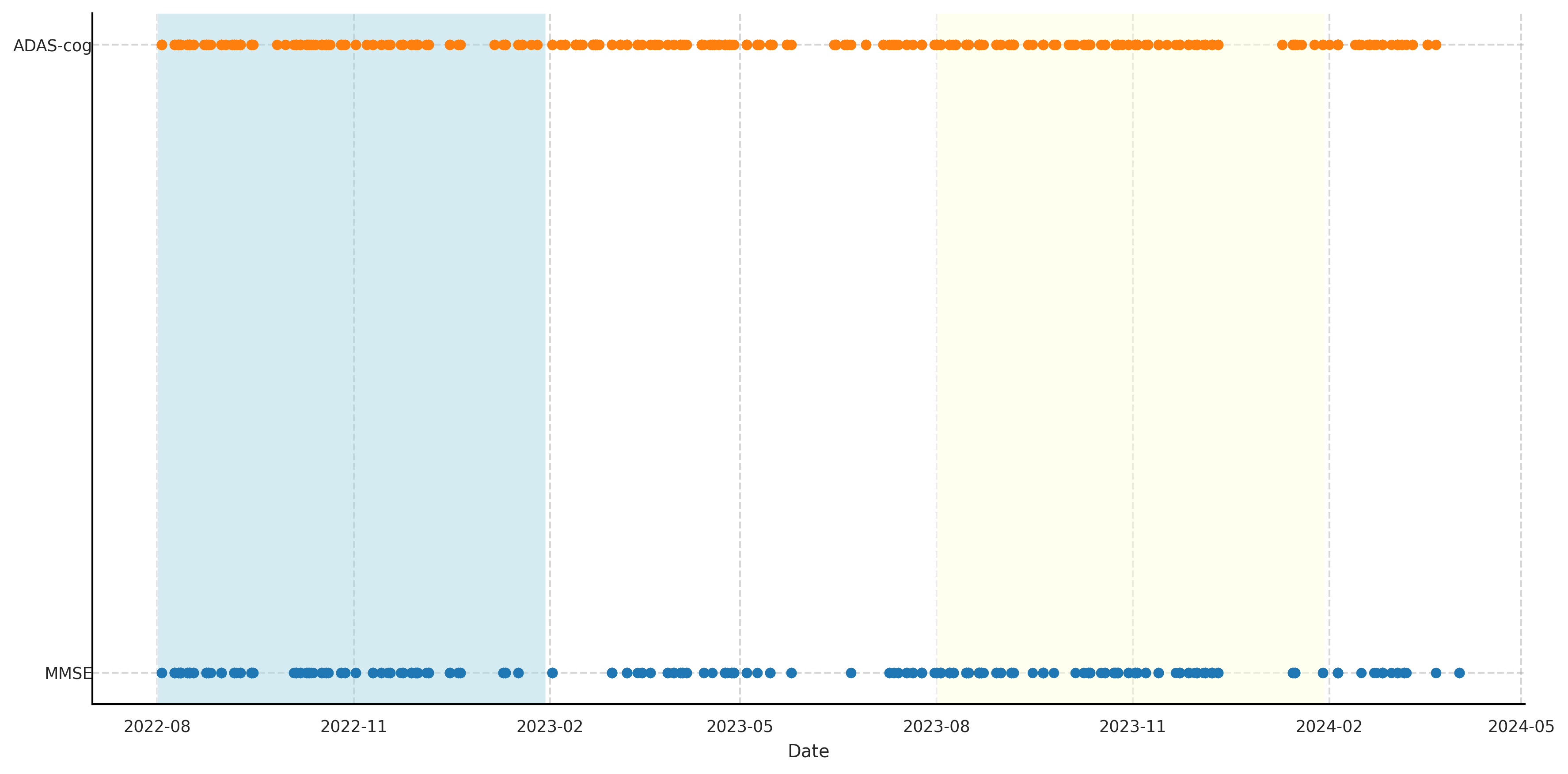}
    \caption{Timeseries of cognitive test of the test set participants}
    \label{cognitive}
\end{figure}

In addition to activity data, we had access to diagnostic information for the 134 participants, including birthdate, gender, living situation (whether they lived alone), ethnicity, and dementia diagnosis. Cognitive assessment scores, such as MMSE and ADAS-Cog, along with their yearly changes, were also available, as is shown in Figure \ref{cognitive}. To mitigate the risk of time series data contamination, the period from July 1, 2021, to July 1, 2023, was used for fine-tuning, while data from July 31, 2023, to January 30, 2024, was reserved for testing. The test set data was used to directly extract coding vectors from the fine-tuned language model and analyze behavioral transfer patterns. However, participants with incomplete records, particularly those with gaps in data after July 2023, were excluded from the test set. The final test set comprised the fifty participants with the most complete data post-July 2023, resulting in 869 comprehensive clinical records. The cognitive test results of participants in the test set, from August 2023 through the cut-off date of January 30, 2024, were used for our analyses. Additionally, we incorporated their cognitive test results from one year earlier to assess changes in MMSE and ADAS-Cog scores over time.

\subsection{Detailed Training Process}
\label{append_training}

Since the sensor data are recorded with second-level precision, each participant generates 86,400 data points per day, far exceeding the input token limit of the language model we are using (which allows a maximum of 256 tokens). To address this, the raw data were downsampled by extracting discrete values at 20-minute intervals, reducing the data points per day to 72. After converting these data points to strings, the token count is 72, which falls within the model's token limit.

Given the unlabeled nature of our temporal data, we employ a cluster-based contrastive sample selection approach for model training. This method leverages the inherent structure within the data to create meaningful positive and negative sample pairs. The detailed steps are as follows:

\begin{enumerate}
    \item \textbf{One-hot Encoding:} Convert all daily string representations into one-hot encoded vectors.
    \item \textbf{Clustering:} Apply K-means clustering to the one-hot encoded vectors to group similar daily patterns into clusters. 
    \item \textbf{Target Day Selection:} Choose a specific day as the target for comparison.
    \item \textbf{Similar Sample Selection:} For the target day, select a similar sample that meets all the following criteria:
        \begin{itemize}
            \item From the same participant
            \item Within a 30-day window of the target day
            \item Belongs to the same cluster as the target day
        \end{itemize}
    \item \textbf{Dissimilar Sample Selection:} Randomly select any other sample that does not meet the criteria for similar sample selection.
\end{enumerate}

We selected a 30-day interval for positive sample selection for two key reasons: first, k-means clustering of the encoded vectors yielded the best results with a 30-day window, as is shown in \ref{ablation}; second, many patients undergo regular physical checkups, such as urine tests, on a monthly basis, aligning well with this time frame.

\begin{table}[htbp]
\label{ablation}
\centering
\caption{Silhouette scores under different models and parameter settings}
\begin{tabular}{l@{\hskip0.2in}r@{\hskip0.4in}r@{\hskip0.2in}r@{\hskip0.2in}r@{\hskip0.2in}r@{\hskip0.2in}}
\toprule
Model & Parameters & \multicolumn{4}{c}{Silhouette scores} \\
\cmidrule(r){3-6}
& & 4 & 5 & 6 & 7 \\
\midrule
MiniLM- & 7days & 0.459 & 0.451 & 0.431 & 0.413 \\
L12-v2  & 30days & \bfseries 0.554 & \bfseries 0.554 & \bfseries 0.554 &  \bfseries 0.542 \\
& 180days & 0.437 & 0.429 & 0.370 & 0.407 \\
\midrule
BAAI/bge- & 30days & 0.459 & 0.425 & 0.473 & 0.473 \\
small-v1.5\citep{xiao_c-pack_2023} & no tune & 0.173 & 0.165 & 0.181 & 0.170 \\
\bottomrule
\end{tabular}
\end{table}

This ablation study is conducted to address potential concerns with our initial assumptions and to select the most suitable parameters that maximize the separation of different vector embeddings. By examining the k-means clustering results under various settings, we aim to identify the optimal configuration that yields the greatest distinction among the vector representations, mitigating potential issues arising from our assumptions.

In each epoch, we randomly selected 50,000 triplets from the dataset, using a batch size of 256. Sentence embeddings were evaluated using a triplet loss, where the Manhattan distance was calculated and optimized between the coding vectors of anchor samples and their corresponding positive and negative samples. Manhattan distance's suitability for sparse data, computational efficiency, and applicability in discrete systems make it a preferred choice for measuring similarity in our work. The loss function was optimized using the AdamW algorithm \citep{loshchilov_decoupled_2019} with a learning rate of $2 \times 10^{-5}$ and a weight decay of 0.01. A linear warm-up learning rate scheduler was applied with 10,000 warm-up steps. The training of large language model was carried out using NVIDIA A100 GPU, with each training epoch taking approximately 444.19 seconds.

\subsection{PageRank Model for a Single Patient}
\label{pagerank}
\subsubsection{Model Definition}

We aim to compute the PageRank model fit and entropy value for a single patient based on their embeddings and cluster labels. The process involves defining a transition matrix based on distances between clusters, computing the PageRank scores.

\subsubsection{Transition Matrix Construction}

Given:
\begin{itemize}
    \item $X$: Patient embeddings (shape: $(n\_samples, 2)$).
    \item $y$: Patient cluster labels (shape: $(n\_samples,)$).
    \item $num\_clusters$: Number of clusters.
    \item $threshold$: Distance threshold for defining transitions.
\end{itemize}

The transition matrix $\mathbf{T}$ is computed as follows:

\begin{equation}
T_{ij} = \frac{\sum_{k \in C_i} \sum_{l \in C_j} \mathbb{1}_{\{d(k, l) \leq \text{threshold}\}}}{\sum_{l \in C_i} \sum_{m \in C_j} \mathbb{1}_{\{d(l, m) \leq \text{threshold}\}}}
\end{equation}

where:
\begin{itemize}
    \item $C_i$ and $C_j$ are the sets of samples in clusters $i$ and $j$, respectively.
    \item $d(k, l)$ denotes the distance between samples $k$ and $l$.
    \item $\mathbb{1}_{\{ \cdot \}}$ is an indicator function that equals 1 if the condition is true and 0 otherwise.
\end{itemize}

\subsubsection{PageRank Computation}

The PageRank vector $\mathbf{p}$ is computed iteratively using:

\begin{equation}
\mathbf{p}^{(t+1)} = \frac{1 - \alpha}{num\_clusters} + \alpha \mathbf{T}^\top \mathbf{p}^{(t)}
\end{equation}

where $\alpha$ is the damping factor (typically 0.85), and $\mathbf{T}^\top$ is the transpose of the transition matrix. The process continues until convergence:

\begin{equation}
\| \mathbf{p}^{(t+1)} - \mathbf{p}^{(t)} \|_1 < \text{tol}
\end{equation}

where $\text{tol}$ is a predefined tolerance for convergence.

\subsubsection{Algorithm Summary}
Algorithm~\ref{algorithm} explains the specific implementation steps of the PageRank we use. 

\begin{algorithm}
\caption{PageRank for a Single Patient}
\begin{algorithmic}
\STATE \textbf{Input:} $X$, $y$, $num\_clusters$, $threshold$, $\alpha$, $\text{max\_iter}$, $\text{tol}$
\STATE \textbf{Output:} $\mathbf{T}$, $\mathbf{p}$
\STATE Initialize transition matrix $\mathbf{T}$ with zeros
\FOR{each cluster $i$}
    \FOR{each cluster $j$}
        \STATE Compute distances between samples in clusters $i$ and $j$
        \STATE Update $\mathbf{T}_{ij}$ based on the distance threshold
    \ENDFOR
\ENDFOR
\STATE Normalize transition matrix $\mathbf{T}$
\STATE Initialize PageRank vector $\mathbf{p}$ uniformly
\FOR{iteration $t = 1$ to $\text{max\_iter}$}
    \STATE Compute new PageRank vector $\mathbf{p}^{(t+1)}$
    \IF{convergence condition met}
        \STATE Break
    \ENDIF
\ENDFOR
\STATE Compute PageRank matrix $\mathbf{P}_{\text{rank}}$
\end{algorithmic}
\label{algorithm}
\end{algorithm}

\subsubsection{Algorithm Visualization}
\label{pagerank_vis}

\begin{figure}[ht]
    \centering
    \subfigure{
        \includegraphics[width=0.4\textwidth]{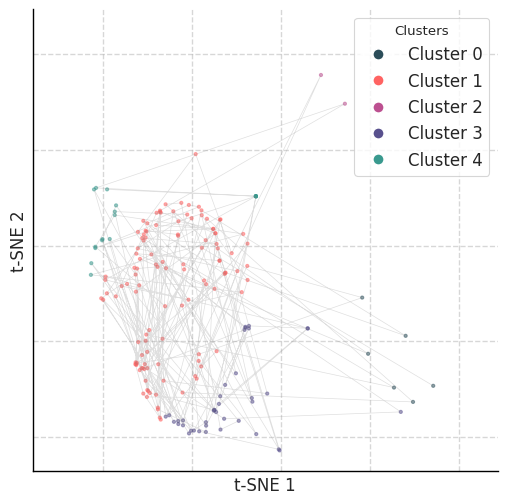}
        \label{fig:fig1}
    }
    \hfill
    \subfigure{
        \includegraphics[width=0.4\textwidth]{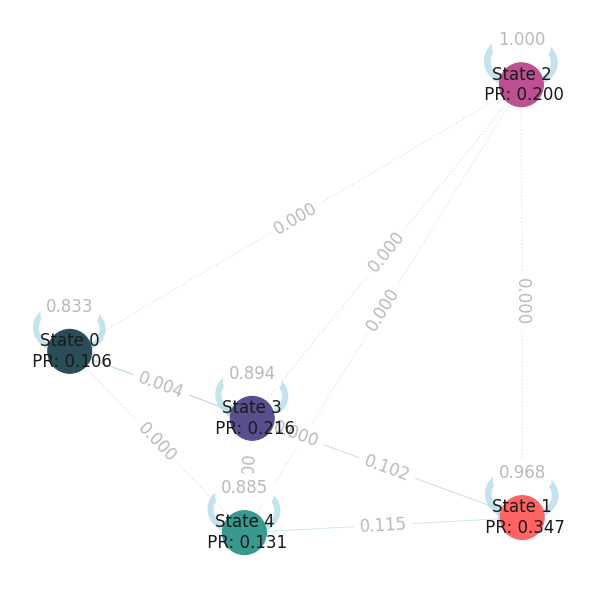}
        \label{fig:fig2}
    }
    \caption{Visualization of the generation of Pagerank value, left graph is single participant 2D t-SNE visualization, right graph is PageRank nodes value visualization}
    \label{fig:figures}
\end{figure}

Figure \ref{fig:figures} is a sample PageRank state generation process.

\subsection{T-SNE plots for test set}
\begin{figure}[htbp]
    \centering
    \includegraphics[width=0.8\textwidth]{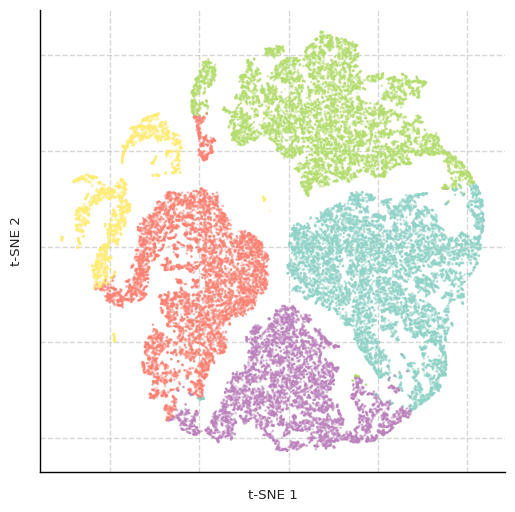}
    \caption{T-SNE for embedded datapoints in the test set}
    \label{tsneplotpic1}
\end{figure}

Figure \ref{tsneplotpic1} shows the best clustering result using our encoding language model.
\label{tsneplot1}

\subsection{T-SNE plots for individual participants in test set}
\label{tsneplot2}

\begin{figure}[htbp]
    \centering
    \includegraphics[width=0.8\textwidth]{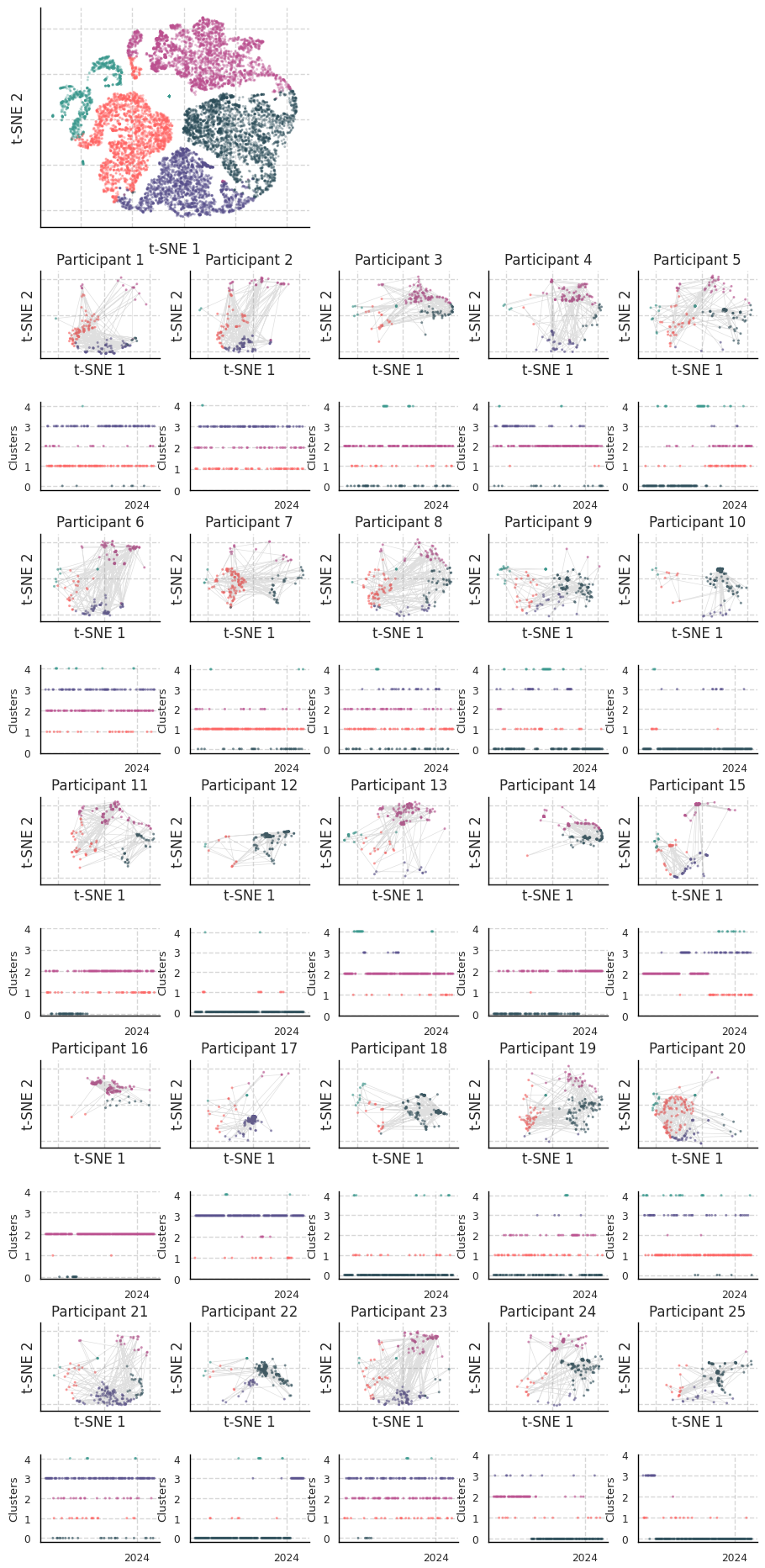}
    \caption{T-SNE for individuals in test set}
    \label{tsneplotpic2-1}
\end{figure}

\begin{figure}[htbp]
    \centering
    \includegraphics[width=0.8\textwidth]{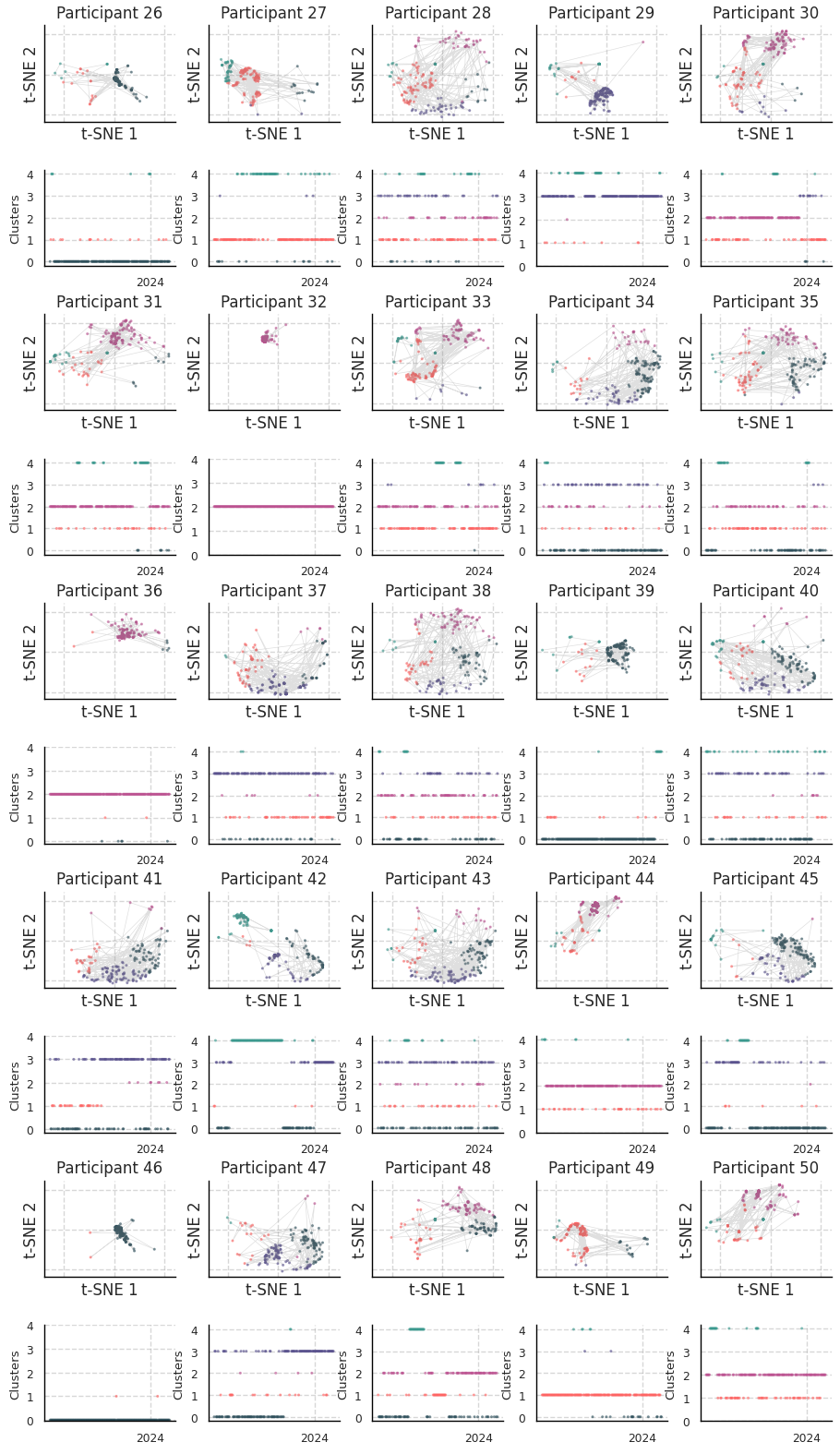}
    \caption{T-SNE for individuals in test set}
    \label{tsneplotpic2-2}
\end{figure}

Figure \ref{tsneplotpic2-1} and Figure \ref{tsneplotpic2-2} illustrate individual movement in embedded space, with their location and timespan. Every datapoint is collected after 2023-07-31.

\subsection{Relations between most similar and dissimilar pariticipants}
\label{similar}
This section summarizes the analysis of patient characteristics based on similarity metrics calculated from PageRank state embeddings. The aim is to compare characteristics between selected participants, in this case using all 50 participants, their three most similar counterparts, and their least similar counterparts. The rate of change in MMSE and ADAS-Cog scores ($\Delta MMSE$ and $\Delta ADAS-Cog$) was calculated by subtracting the participants' scores from one year prior from their current scores within the selected time period, and normalizing the result by the time difference .

\begin{table}[htbp]
\caption{P-values for characteristics comparison with most similar patients.}
    \centering
    \begin{tabular}{lcc}
        \toprule
        \textbf{Feature} & \textbf{P-Value} (n=50) \\
        \midrule
        MMSE & 0.9978 \\
        ADAS-Cog & 0.9788 \\
        HADS - Depression Score & 0.8171 \\
        HADS - Anxiety score & 0.8239 \\
        Age & 0.8860 \\
        $\Delta\ MMSE$  & 0.3773 \\
        $\Delta\ ADAS-Cog$ & 0.5628 \\
        \bottomrule
    \label{p1}
    \end{tabular}
    
    \label{tab:p_values_most_similar}
\end{table}

\begin{table}[htbp]
\caption{P-values for characteristics comparison with least similar patients.}
    \centering
    \begin{tabular}{lcc}
        \toprule
        \textbf{Feature} & \textbf{P-Value}  (n=50) \\
        \midrule
        MMSE & 0.8969 \\
        ADAS-Cog & 0.8696 \\
        HADS - Depression Score & 0.2560 \\
        HADS - Anxiety score & \bfseries 0.0069 \\
        Age & \bfseries 4.5934e-05 \\
        $\Delta\ MMSE$ & \bfseries 0.0224 \\
        $\Delta\ ADAS-Cog$ & 0.4426 \\
        \bottomrule
    \end{tabular}
   
    \label{tab:p_values_least_similar}
\end{table}

\begin{table}[htbp]
\caption{Effect sizes for comparisons with most similar patients.}
    \centering
    \begin{tabular}{lcc}
        \toprule
        \textbf{Feature} & \textbf{Effect Size (Cohen's d)} \\
        \midrule
        MMSE & -1.3979 \\
        ADAS-Cog & 2.3360 \\
        HADS - Depression Score & 1.2452 \\
        HADS - Anxiety score & 2.7198 \\
        Age & 0.8310 \\
        $\Delta\ MMSE$ & -0.0999 \\
        $\Delta\ ADAS-Cog$ & -0.7578 \\
        \bottomrule
    \end{tabular}
    
    \label{tab:effect_sizes_most_similar}
\end{table}

\begin{table}[htbp]
\caption{Effect sizes for comparisons with least similar patients.}
    \centering
    \begin{tabular}{lcc}
        \toprule
        \textbf{Feature} & \textbf{Effect Size (Cohen's d)} \\
        \midrule
        MMSE & -0.9228 \\
        ADAS-Cog & 3.1013 \\
        HADS - Depression Score & -2.9330 \\
        HADS - Anxiety score & -6.6526 \\
        Age & -24.8634 \\
        $\Delta\ MMSE$ & -23.0227 \\
        $\Delta\ ADAS-Cog$ & 5.7778 \\
        \bottomrule
    \end{tabular}
    \label{tab:effect_sizes_least_similar}
\end{table}

Table \ref{tab:p_values_most_similar}, \ref{tab:p_values_least_similar}, \ref{tab:effect_sizes_most_similar}, \ref{tab:effect_sizes_least_similar} indicates that there are various degrees of differences between selected patients and their most and least similar counterparts. The p-values suggest that for most features, there are no statistically significant differences between the selected patients and their most similar or least similar counterparts, except for the HADS - Anxiety score Age and change in Adas-cog scores in the least similar group. The effect sizes provide insight into the magnitude of differences, where large effect sizes are observed in some features such as HADS - Anxiety score Age and change in Adas-Cog scores.

\subsection{MMSE vs ADAS-Cog Scores Scatter Plot by Pagerank vector Clustering}
\label{clustering}

\begin{figure}[htbp]
    \centering
    \includegraphics[width=0.8\textwidth]{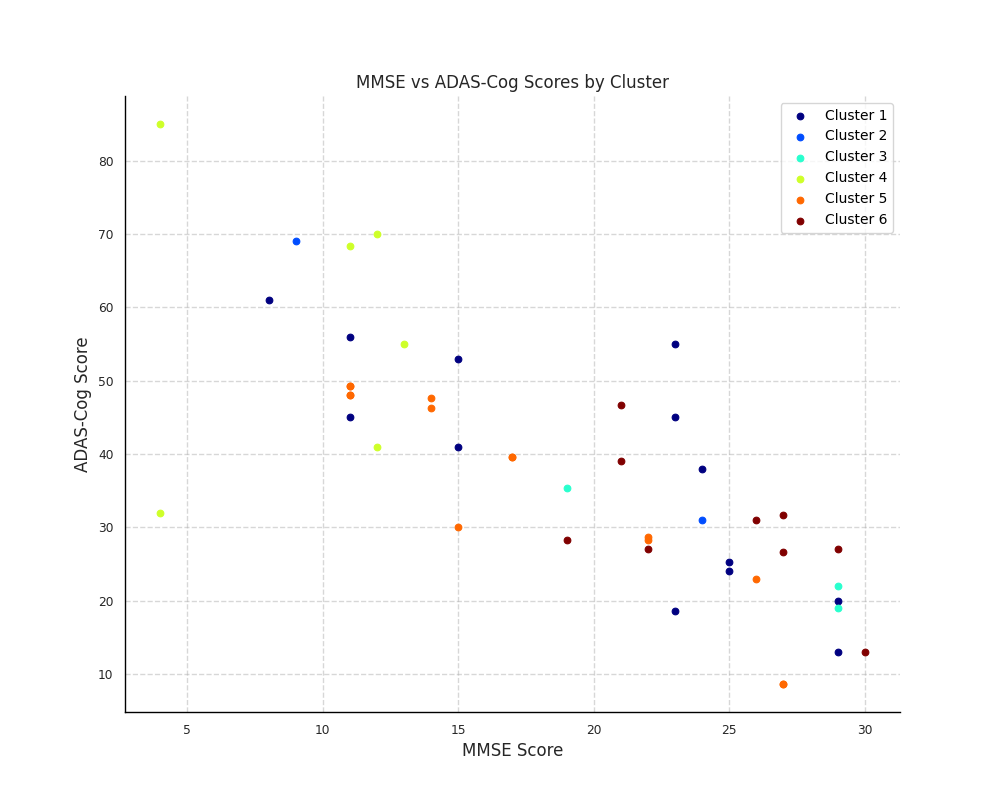}
    \caption{MMSE vs ADAS-Cog Scores by Cluster}
    \label{scatter}
\end{figure}

Scatter plot Figure \ref{scatter} illustrates the relationship between the MMSE Score (Mini-Mental State Examination) on the x-axis and the ADAS-Cog Score (Alzheimer’s Disease Assessment Scale-Cognitive Subscale) on the y-axis. The plot is color-coded based on six distinct clusters, which has the best silhouette score in K-means Clustering for PageRank vector.

\subsubsection{Key Observations}
\textbf{Inverse Relationship}: The ADAS-Cog score, which measures cognitive impairment, tends to decrease as the MMSE score increases. A higher MMSE score indicates better cognitive function, and correspondingly, a lower ADAS-Cog score implies less cognitive impairment.

\subsubsection{Cluster Distribution}
\begin{itemize}
    \item \textbf{Cluster 1 (Dark Blue)}: Data points are spread across a wide range of ADAS-Cog scores from 50 to 80 and correspond to MMSE scores between 5 and 25. This cluster likely represents individuals with higher cognitive impairment.
    
    \item \textbf{Cluster 2 (Light Blue)}: Data points are mainly grouped between MMSE scores of 20 and 30, with ADAS-Cog scores ranging between 40 and 70.
    
    \item \textbf{Cluster 3 (Green)}: This cluster includes individuals with intermediate MMSE and ADAS-Cog scores, generally between 10 to 25 on the MMSE scale and 30 to 60 on the ADAS-Cog scale.
    
    \item \textbf{Cluster 4 (Yellow)}: Represented sparsely with fewer points, indicating individuals with higher ADAS-Cog scores (around 50–80) and lower MMSE scores (10–15).
    
    \item \textbf{Cluster 5 (Orange)} and \textbf{Cluster 6 (Red)}: These clusters consist of individuals with generally higher MMSE scores (15 to 30) and lower ADAS-Cog scores, signifying lesser cognitive impairment.
\end{itemize}

Overall, the plot provides a clear visualization of cognitive function across individuals, with each cluster highlighting groupings based on MMSE and ADAS-Cog scores. The inverse trend between these two cognitive measures is evident, offering insights into patterns of cognitive decline.

\end{document}